\documentclass[runningheads]{llncs}
\usepackage{graphicx}
\usepackage[utf8]{inputenc}
\usepackage{float}
\usepackage{amsmath}
\usepackage{textcomp}
\usepackage{booktabs}
\usepackage{xcolor}

\graphicspath{{figs/}} \DeclareGraphicsExtensions{.png}

\begin{document}

\title{Efficient embedding network for 3D brain tumor segmentation}
%
%

\author{Hicham Messaoudi\inst{1} \and Ahror Belaid\inst{1} \and  Mohamed Lamine Allaoui\inst{1}
\and Ahcene Zetout\inst{1}  \and Mohand Said Allili \inst{2} \and
Souhil Tliba\inst{1,3} \and Douraied~Ben Salem\inst{4,5}
\and~Pierre-Henri Conze\inst{4,6}}
%
%
\institute{Medical Computing Laboratory (LIMED), University of Abderrahmane Mira,\\
              06000, Bejaia, Algeria, \url{http://http://www.univ-bejaia.dz/limed/}\\
              \email{ahror.belaid@univ-bejaia.dz}\\
 \and
Universit\'{e} du Qu\'{e}bec en Outaouais, Gatineau, Qu\'{e}bec, J8X 3X7
 \and
 Neurosurgery Department, University Hospital Center, \\
  Biological Engineering of Cancers, 06000 Bejaia, Algeria
 \and
 Laboratory of Medical Information Processing (LaTIM)\\
               UMR 1101, Inserm, 22 avenue Camille Desmoulins, 29238 Brest, France\\
\and
            Neuroradiology Department, CHRU la cavale blanche,\\
            Boulevard Tanguy-Prigent, 29609 Brest, France\\
\and
            IMT Atlantique, Technopôle Brest Iroise, 29238 Brest , France\\
%
             }

\authorrunning{A. Belaid et al.}
\maketitle              
\begin{abstract}
3D medical image processing with deep learning greatly suffers from
a lack of data. Thus, studies carried out in this field are limited
compared to works related to 2D natural image analysis, where very large
datasets exist. As a result, powerful and efficient 2D
convolutional neural networks have been developed and trained. In
this paper, we investigate a way to transfer the performance of a
two-dimensional classification network for the purpose of
three-dimensional semantic segmentation of brain tumors. We propose
an asymmetric U-Net network by incorporating the EfficientNet model
as part of the encoding branch. As the input data is in 3D, the first layers
of the encoder are devoted to the reduction of the third dimension
in order to fit the input of the EfficientNet network.
Experimental results on validation and test data from the BraTS 2020
challenge demonstrate that the proposed method achieve promising
performance.



\end{abstract}
\section{Introduction}

Gliomas are the most common type of primary brain tumors of the
central nervous system. They can be of low-grade or
high-grade type. High-Grade Gliomas (HGG) are an aggressive type of malignant
brain tumors that grow rapidly. Furthermore, Low-Grade Gliomas (LGG) are
classified into grade I and grade II. These tumors
represent less than 50\% of glial tumors and are considered as
isolated tumor cells within the nervous parenchyma, with slow
initial growth. Gliomas are also characterized by their infiltrating
character resulting in ambiguous and fuzzy boundaries.

However, the treatment of LGG remains difficult due to the
variability of tumor size, location, histology and biological
behavior. Furthermore, because of the pressure that the tumor
exhibits, normal tissues get deformed, making it even harder to
distinguish normal tissues from tumoral areas. The diagnosis of
these tumors followed by early treatments are critical for patient
survival. Indeed, in most cases, patients who are suffering from LGG
die in the next ten years of the initial diagnosis. For all these
reasons, accurate and reproducible segmentation of gliomas is a
pre-requisite step for investigating brain MRI data.

Magnetic Resonance Imaging (MRI) has been quickly imposed as being
an essential medical imaging modality for disease diagnosis. MRI is
particularly useful for brain tumor diagnosis, patient follow-up,
therapy evaluation and human brain
mapping~\cite{Bauer13,Zaouche18,Belaid20}. The main advantage
related to the use of MRI is its ability of acquiring non-invasive
and non-irradiant medical images. It is also very sensitive to the
contrast and provides an excellent spatial resolution which is
entirely appropriate for the exploration of the brain tissues
nature. In addition, the imaging easily derives 3D volumes according
to brain tissues.

The multimodal Brain Tumor Segmentation (BraTS)
challenge~\cite{Menze15,Bakas17,Bakas17a,Bakas17b,Bakas18} aims at
encouraging the development of state-of-the-art methods for the
segmentation of brain tumors by providing a large 3D MRI dataset of
annotated LGG and HGG. The BraTS 2020 training dataset include 369
cases (293 HGG and 76 LGG), each with 4 modalities describing:
native (T1), post-contrast T1-weighted (T1Gd), T2-weighted (T2), and
T2 Fluid Attenuated Inversion Recovery (T2-FLAIR) volumes, which
were acquired with different clinical protocols and various MRI
scanners from multiple (n=19) institutions. Each tumor was segmented
into edema, necrosis and non-enhancing tumor, and active/enhancing
tumor. Annotations were combined into 3 nested sub-regions: Whole
Tumor (WT), Tumor Core (TC) and Enhancing Tumor (ET).

In the last decade, Convolutional Neural Networks (CNNs) have
outperformed all others traditional methodologies in biomedical
image segmentation. Particularly, the U-Net
architecture~\cite{Ronneberger15} is currently experiencing a huge
success, as most winning contributions to recent medical image
segmentation challenges were exclusively build around U-Net.

The aim of this work is to investigate how to re-use powerful deep
convolutional networks that exist in the literature for 2D image
analysis purposes. Indeed, a lot of powerful 2D classification
networks are well trained on very large datasets. Through transfer
learning, these pre-trained networks can be easily re-used for other
classification problems.
However, transferring both learning and feature detection power of
classification networks to another types of problem is not obvious.
Especially, in problems using convolutional encoder-decoder
architectures such as U-Net, this consists in integrating the
pre-trained model as part of the encoder branch. When the dimensions of the
classifier and the processed network are the same (i.e. when they
processed data having the same dimensions), the integration of the
classifier is almost immediate. Otherwise, an adaptation process is
necessary to be able to adapt the classifier to deal with the dimensions of the images under study.

Related works can be found in the literature  where the segmentation
of 3D medical images are based on 2D networks whose encoder has been
pre-trained on ImageNet~\cite{Conze20,Conze20_,Vu19}. Although
taking advantage of a pre-trained encoder, these networks do not
integrate the spatial coherence in 3D, since they process data slice
by slice. In our case, we aim at studying how to transfer the skills
of a powerful and pre-trained network on 2D images to a
convolutional encoder-decoder processing 3D images without loosing
the consistency across the third dimension. It is known that the
field of 2D image classification is well studied and well developed,
generally because of the availability of large annotated 2D natural
image datasets. The public availability of such large 3D databases
is non-existent, and even less so in the medical field. Taking
advantage of such powerful and well-studied networks by re-using
them in a 3D problem is an idea worth investigating.

In summary, the main proposed contributions are :
\begin{itemize}
  \item Proposal of an efficient way to transfer any 2D classification
architecture to 3D segmentation purposes without losing the 3D
consistency.
  \item The proposed idea is generalizable in order to integrate any
low-dimensional classification architecture into another
high-dimensional architecture without losing spatial coherence.
\end{itemize}

In particular, we are interested in the recently published
EfficientNet network~\cite{Tan19} which achieves state-of-the-art
top-1 and top-5 accuracy on ImageNet, while being widely more
smaller and faster on inference than the best existing deep
architectures. This paper is organized as follow: the proposed
architecture is detailed in the next section (2), the results are
summarized in section 3, followed by a conclusion in section 4.

\section{Method}

The proposed segmentation approach follows a convolutional
encoder-decoder architecture. It is built from an asymmetrically
large encoder to extract image features and a smaller decoder to
reconstruct segmentation masks. We embed as part of the encoder
branch the recently proposed network called
EfficientNet~\cite{Tan19}.

\begin{figure}
\includegraphics[width=\textwidth]{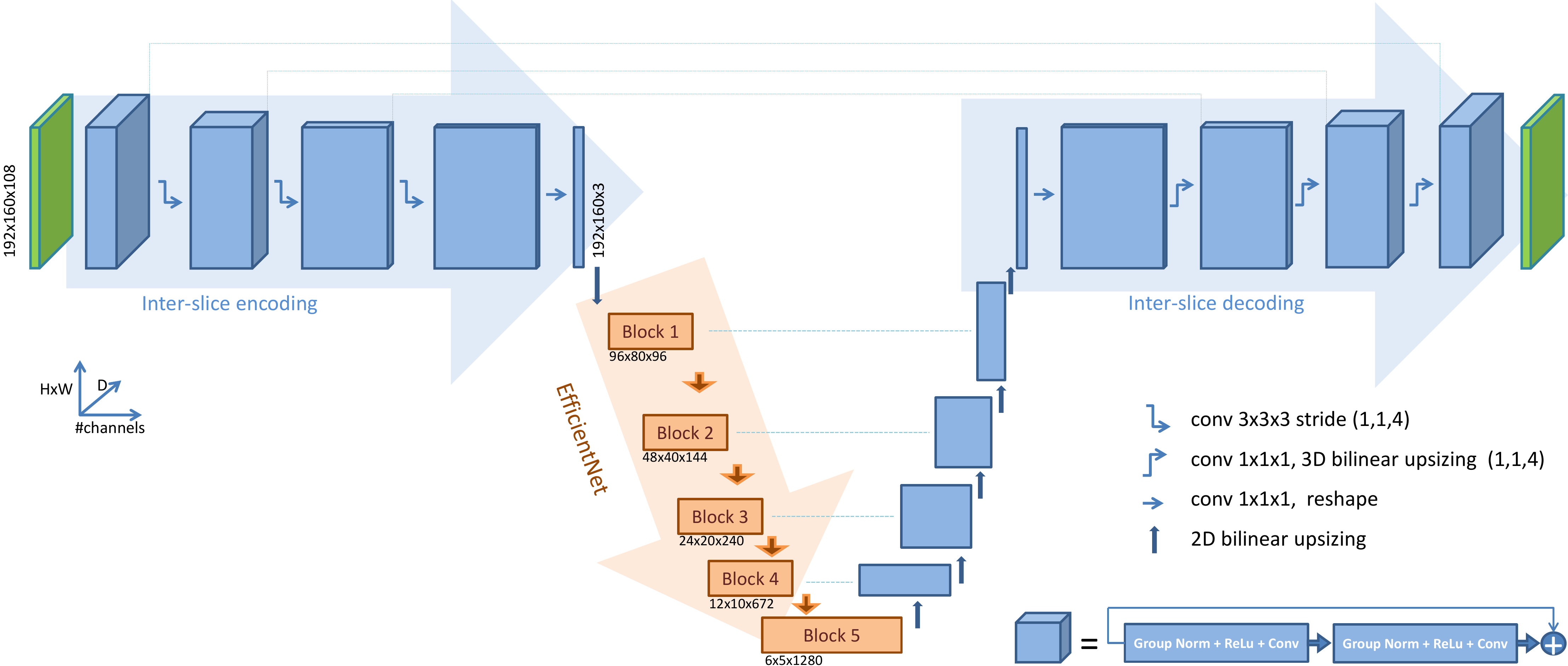}
\caption{Schematic illustration of the proposed network
architecture. Input is a one-channel cropped 3D MRI. The inter-slice
encoder as well as the decoder consist of a succession of residual
blocks with GroupNorm normalization. The output of the decoder has
three channels with the same spatial size as the input. Under each
EfficientNet block are shown corresponding output feature
dimensions.} \label{fig:Schematic}
\end{figure}

\subsection{Data pre-processing}

Because of the limitations in GPU memory and time-consuming
computation, we were forced to take some precautions. We process
each modality separately, and resize the images dimensions
by reducing the background using the largest crop size of
$192\times160\times108$, and compromise the batch size to be 1. We
do not use any additional training data and employ the provided
training set only. We normalize all input images to have zero mean and
unit variance.

Since the EfficientNet models range from 0 to 7, we restricted our
experimentations in this preliminary work only by testing the
baseline EfficientNet-B0~\cite{Tan19}. Indeed, this model presents a compromise
between performance and complexity. Using EfficientNet-B7 at this
stage requires more resources, a choice which may strongly complexify our
model. In any case, if the proposed architecture works well with
this baseline model, its generalization towards EfficientNet-B7,
will be straightforward and will definitely improve the performances.

\subsection{Encoder branch}

The encoding process goes through two steps.
First, we encode three-dimensional data into two-dimensional data,
while keeping the height and width at their original size and
compressing only the depth to 3 channels. Second, the data is now
ready to start the second encoding step, which is none other than
the EfficientNet network without its fully connected layers.

As shown in Fig.~\ref{fig:Schematic}, EfficientNet is represented as
blocks as in its original version. However, only the blocks involved
in skip connection layers are represented. The inter-slices encoding
part uses convolutional blocks which consists of two convolutional
layers with normalization and ReLU, followed by skip connection.
Following the works of~\cite{Myronenko18}, we choose to use Group
Normalization~\cite{Wu18}, which divides the channels into groups
before normalizing them by using mean and variance of each group. It
seems to perform better than traditional batch normalization,
especially when the batch size is small.

Let us assume that the input volume is of width W, height H, depth D,
with C channels. The data passes through a 3D-2D shrinking step.
Thus, the depth is reduced by factors 3, 3 and 4 to reach a final
depth size of 3, which corresponds to the required number of
channels of EfficientNet. In this shrinking procedure, the width and
height of the single 3D batch are not modified. The different depth reduction factors can be changed and adapted
according to the third dimension of the data. In our study, dimensions
change as shown in Tab.~\ref{tab:dim}. After this process, the
reduced data reached through the 2D shrinking process are given as inputs of the
EfficientNet model.

\begin{center}
\begin{table}[]
\centering
\begin{tabular}{ll}
\hline
Layer &  Dimension \\
\hline
Input &  $W\times H\times D\times C$\\
block 1 &  $W\times H\times \frac{D}{f_1}\times C_1 $\\
block 2 & $ W\times H\times \frac{D}{f_2}\times C_2$\\
\vdots &  $\vdots$\\
block n &  $W\times H\times 3 \times C_n$ \\
block n+1  &  $W\times H\times 3 \times 1 $\\
output   & $ W\times H\times 3 $ \\
\hline \\
\end{tabular}
\centering
\caption{Dimension shrinking.}
\label{tab:dim}
\end{table}
\end{center}

\subsection{Decoder branch}

Asymmetrically to the encoding part, the decoder is composed
entirely of homogeneous blocks as shown in Fig.~\ref{fig:Schematic}.
Obviously, the decoding part linked to the EfficientNet is a 2D
decoder whereas the inter-slice decoding is a 3D decoder. Each
decoder level begins by upsampling the spatial dimension, doubling
the number of features by a factor of 2 followed by skip
connections. A sigmoid function is used as an activation for the output of the decoder which has three channels corresponding to the number of classes with the same spatial size as inputs.\\

\subsection{Loss}

Many networks are trained with a cross-entropy loss function,
however the resulting delineations may not be ideal in terms of Dice
score. As an alternative, one can employ a soft Dice loss function
to train the proposed network. While several formulations of the
dice loss exist in the literature, we prefer to use a soft Dice loss
which has given good results in segmentation challenges in the past
\cite{Myronenko18}. The soft Dice loss function is differentiable
and is given by:
\begin{align}\label{eq:dice}
\mathcal{L}_{Dice}= \frac{2\sum P_{true} P_{pred} }{\sum
P_{true}^2+\sum P_{pred}^2+\epsilon}\enspace,
\end{align}
were $P_{true}$ and $P_{pred}$ represent respectively the ground
truth and the predicted labels. Brain MRI segmentation is a
challenging task partly due to severe class imbalance. Tackling this issue by only using
a fixed loss function, cross entropy or Dice, for the entire
training process is not an optimal strategy. Therefore, a linear
combination of the two loss functions is often considered as the
best practice, and leads to more robust and optimal segmentation
models. In practice, the final loss function is as follows:
\begin{align}\label{eq:loss}
\mathcal{L}=\mathcal{L}_{Cross}-\mathcal{L}_{Dice}\enspace.
\end{align}

\subsection{Training}

The proposed network architecture is trained with centered cropped
data of size $192\times160\times108$ voxels, ensuring that the
useful content of each slice remains within the boundaries of the
cropped area, training was made on BraTS2020~\cite{Menze15} dataset.
Constrained by the poor performance of the material, we set the
batch size to 1. Training has been done using the Adam optimizer
known for little memory requirements, with an initial learning rate
of $10^{-4}$ reduced by a factor of 10 whenever the loss has not
improved for 50 epochs.
\def\wsize#1{\gdef\wsize{#1}}
\wsize{.900}


\section{Results}

We designed the proposed network on Tensorflow and run it on the 368
training cases of BraTS 2020. We process the four modalities
separately and average the sigmoid outputs. The validation dataset
was provided to test the performance of the models on unseen data.
It consists of 125 cases with 4 modalities and without their
corresponding segmentation. The results of the segmentation of the
Whole Tumor (WT), Tumor Core (TC) and Enhancing Tumor (ET) are
summarized in Tab.~\ref{tab:results}. All reported values were
computed by the online evaluation platform
(https://ipp.cbica.upenn.edu/). Fig.~\ref{fig:result2} shows a
typical segmentation results extracted from the validation dataset.

With an average Dice score of $84.13$ for the WT class, on the validation set, the
proposed model seems efficient and accurate enough to handle the
training and inference over a 3D dataset. On the other
hand, results on enhanced and core tumors are less efficient. This
may be due to the fact that we process each MRI modality separately.

After the compression on the depth, the network retains the shape of
the brain and its structure, even if they appear slightly blurred
and degraded. The network learns to extract three axial sections at
more or less regular levels. We notice the presence of the tumor on
the three learned sections, the axial sections generally involved in
3-channel compression are the sections that denote the presence of
tumor information. The high intensity on the tumor parts indicates
that the features of the network are focused on the detection of the
tumor. These details are reported in Fig.~\ref{fig:result3} and
Fig.~\ref{fig:result4}.

The results on the test set are reported on Tab.~\ref{tab:resultstest}.
We can notice a large improvement on ET and TC scores and a slight
decrease on WT compared to the validation set. From
Fig.~\ref{fig:result3}, we can see that the images resulting from
the 3D compression of the axial slices are well oriented on the
learning of the characteristics of the whole tumor, thus the visible
region on FLAIR. In Fig.~\ref{fig:result4}
we find that the compressed data are much more oriented in learning
the structural characteristics of enhanced and necrotic tumor
components, which are the visible regions in T1CE.

\begin{table}[]
\centering
\begin{tabular}{llccc}
 \cmidrule{3-5}
 &&  \multicolumn{3}{c}{Dice} \\
\cmidrule{3-5}
 &&  ET & WT & TC \\
\cmidrule{1-1}\cmidrule{3-5}
Mean && 65.37& 84.13 & 68.04\\
StdDev && 31.93 & 10.67 & 31.29 \\
Median && 81.23 & 87.08 & 78.30 \\
\hline\\
\end{tabular}
\centering \caption{Results on BraTS 2020 validation data. Metrics
were computed by the online evaluation platform.}\label{tab:results}
\end{table}
\begin{table}[] \centering
\begin{tabular}{llccc}
 \cmidrule{3-5}
 &&  \multicolumn{3}{c}{Dice} \\
\cmidrule{3-5}
 &&  ET & WT & TC \\
\cmidrule{1-1}\cmidrule{3-5}
Mean && 69.59& 80.68 & 75.20\\
StdDev && 26.10 & 14.93 & 28.94 \\
Median && 78.81 & 86.38 & 87.28 \\
\hline\\
\end{tabular}
\centering \caption{Results on BraTS 2020 test data. Metrics were
computed by the online evaluation platform.} \label{tab:resultstest}
\end{table}

\def\wsize#1{\gdef\wsize{#1}}
\wsize{.32}

\def\hhsize#1{\gdef\hhsize{#1}}
\hhsize{5cm}
\begin{figure}[]
\begin{minipage}[]{\hsize}
\centering
     \includegraphics[height=\hhsize,width=\wsize\linewidth]{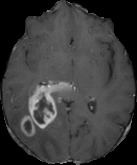}
     \includegraphics[height=\hhsize,width=\wsize\linewidth]{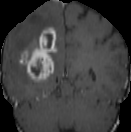}
     \includegraphics[height=\hhsize,width=\wsize\linewidth]{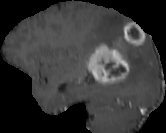}
          \vspace{.21pt}
 \end{minipage}\\
\begin{minipage}[]{\hsize}
\centering
     \includegraphics[height=\hhsize,width=\wsize\linewidth]{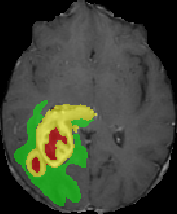}
     \includegraphics[height=\hhsize,width=\wsize\linewidth]{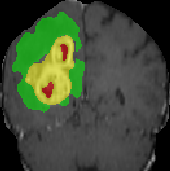}
     \includegraphics[height=\hhsize,width=\wsize\linewidth]{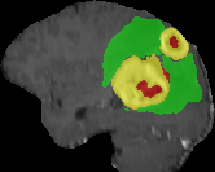}
 \end{minipage}
   \caption{Typical result on the BraTS validation set (2020). From left to right: axial, coronal end sagittal views in
   T1ce. Enhancing tumor is shown in yellow, necrosis in red and edema in green.}
   \label{fig:result2}
\end{figure}

\def\wsize#1{\gdef\wsize{#1}}
\wsize{.32}
\def\hhsize#1{\gdef\hhsize{#1}}
\hhsize{4cm}

\begin{figure}[]
\centering

\begin{minipage}[]{\hsize}
\centering
     \includegraphics[height=\hhsize,width=\wsize\linewidth]{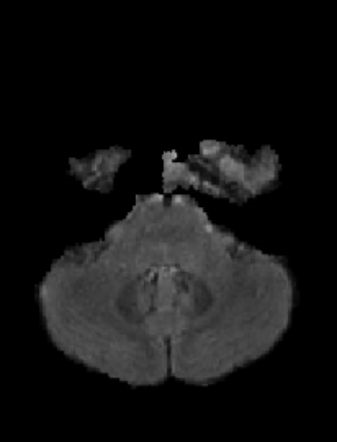}
     \includegraphics[height=\hhsize,width=\wsize\linewidth]{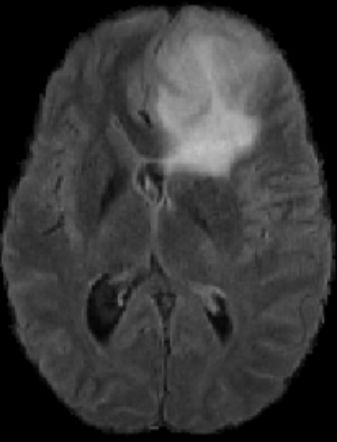}
     \includegraphics[height=\hhsize,width=\wsize\linewidth]{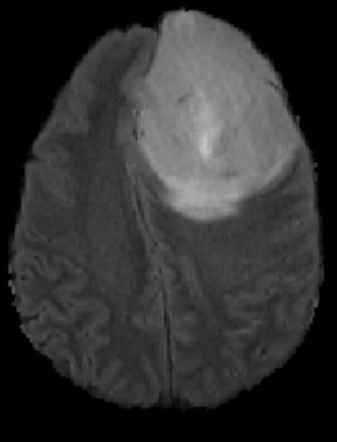}

 \end{minipage}\\
\begin{minipage}[]{\hsize}
\centering
     \includegraphics[height=\hhsize,width=\wsize\linewidth]{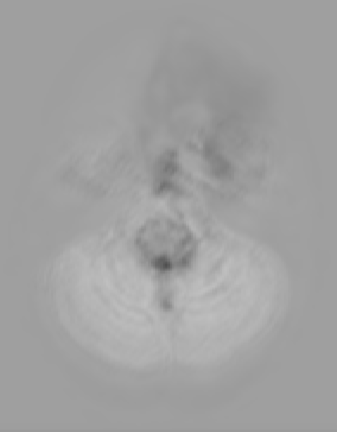}
     \includegraphics[height=\hhsize,width=\wsize\linewidth]{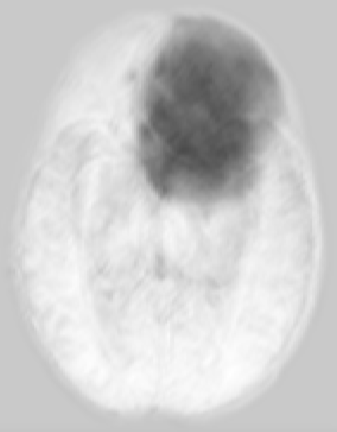}
     \includegraphics[height=\hhsize,width=\wsize\linewidth]{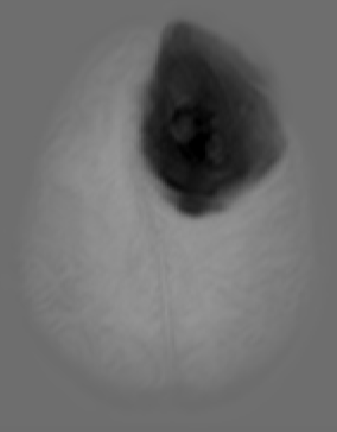}
 \end{minipage}
\begin{minipage}[]{\hsize}
\centering
     \includegraphics[height=\hhsize,width=\wsize\linewidth]{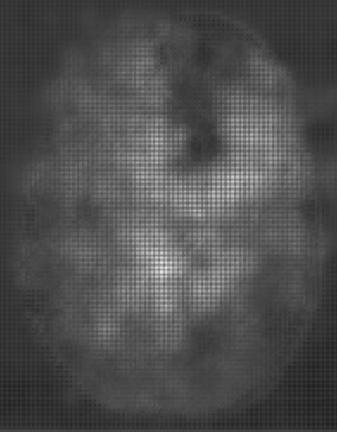}
     \includegraphics[height=\hhsize,width=\wsize\linewidth]{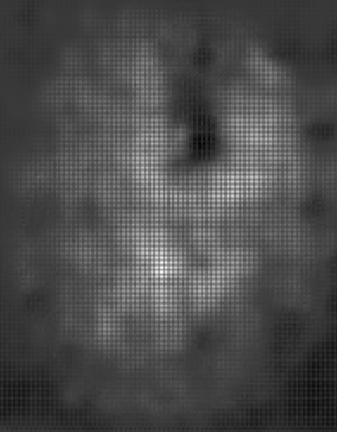}
     \includegraphics[height=\hhsize,width=\wsize\linewidth]{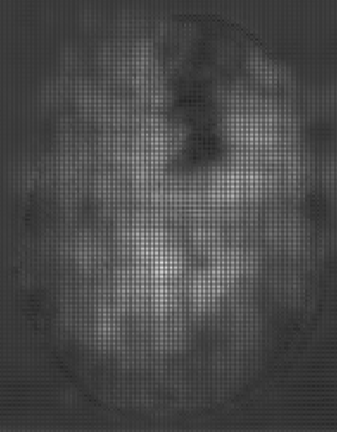}
 \end{minipage}
\begin{minipage}[]{\hsize}
\centering
     \includegraphics[height=\hhsize,width=\wsize\linewidth]{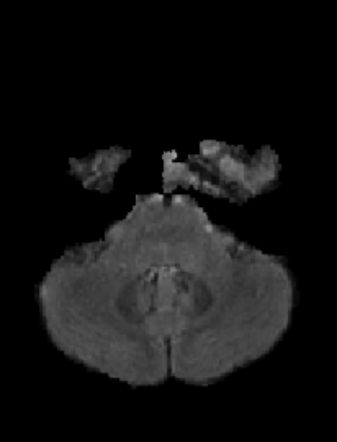}
     \includegraphics[height=\hhsize,width=\wsize\linewidth]{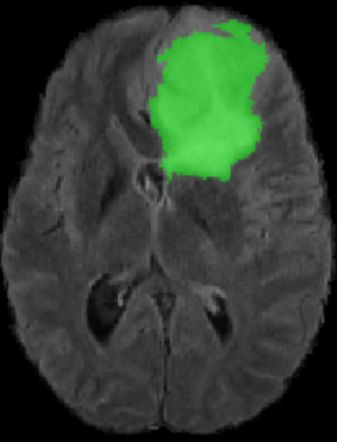}
     \includegraphics[height=\hhsize,width=\wsize\linewidth]{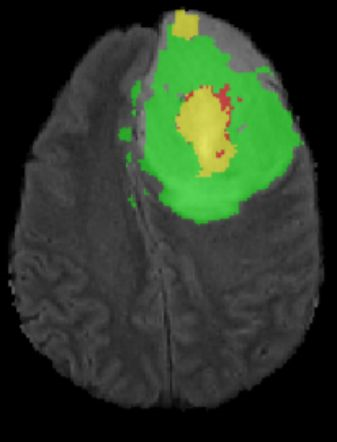}
 \end{minipage}
 \centering
   \caption{Result visualization from the BraTS training set (2020). From up to down : three axial slices of an unique sample in Flair, corresponding channels output of the 3D encoder, corresponding channels input of the 3D decoder and corresponding labels. With Necrotic and Non-Enhancing Tumor core in red, Gadolinium-enhancing tumor in blue and peritumoral edema in green}
   \label{fig:result3}
\end{figure}


\begin{figure}[]
\centering

\begin{minipage}[]{\hsize}
\centering
     \includegraphics[height=\hhsize,width=\wsize\linewidth]{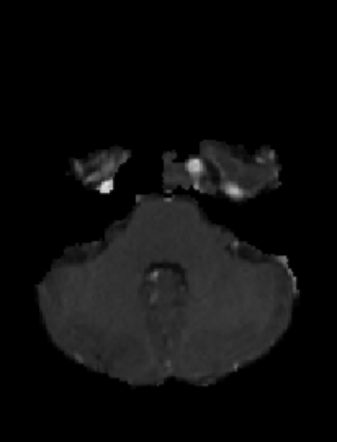}
     \includegraphics[height=\hhsize,width=\wsize\linewidth]{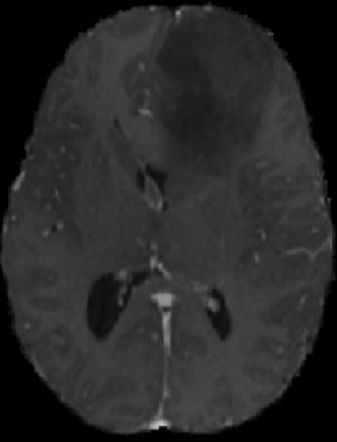}
     \includegraphics[height=\hhsize,width=\wsize\linewidth]{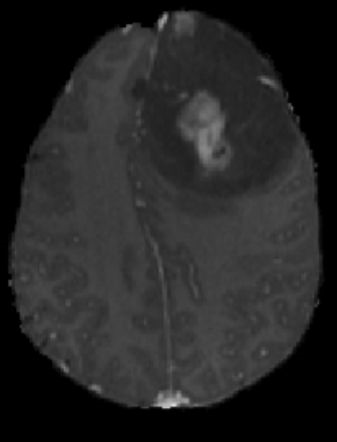}

 \end{minipage}\\
\begin{minipage}[]{\hsize}
\centering
     \includegraphics[height=\hhsize,width=\wsize\linewidth]{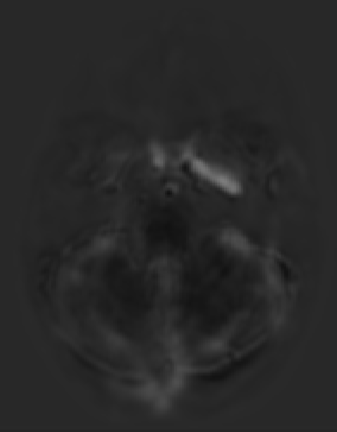}
     \includegraphics[height=\hhsize,width=\wsize\linewidth]{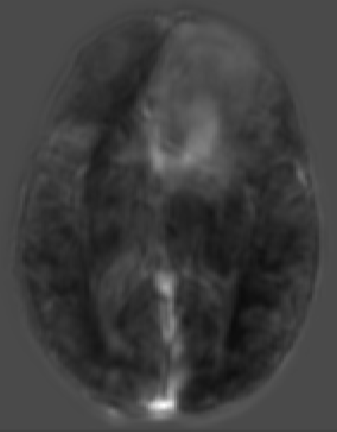}
     \includegraphics[height=\hhsize,width=\wsize\linewidth]{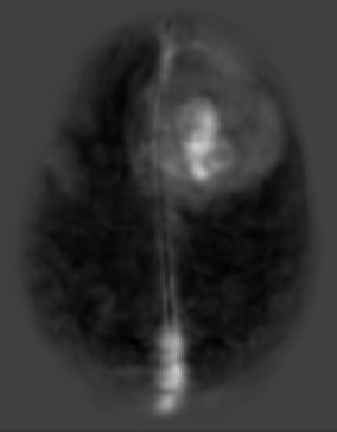}
 \end{minipage}
\begin{minipage}[]{\hsize}
\centering
     \includegraphics[height=\hhsize,width=\wsize\linewidth]{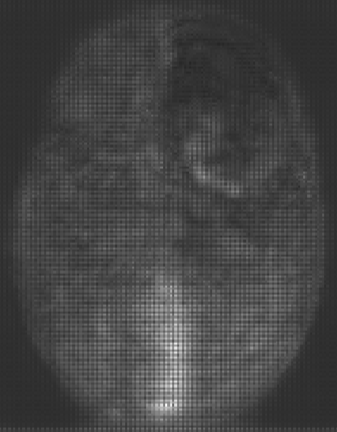}
     \includegraphics[height=\hhsize,width=\wsize\linewidth]{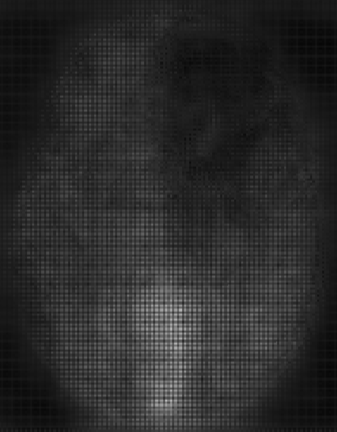}
     \includegraphics[height=\hhsize,width=\wsize\linewidth]{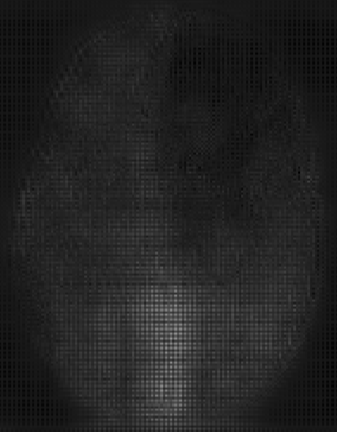}
 \end{minipage}
\begin{minipage}[]{\hsize}
\centering
     \includegraphics[height=\hhsize,width=\wsize\linewidth]{t1ce10}
     \includegraphics[height=\hhsize,width=\wsize\linewidth]{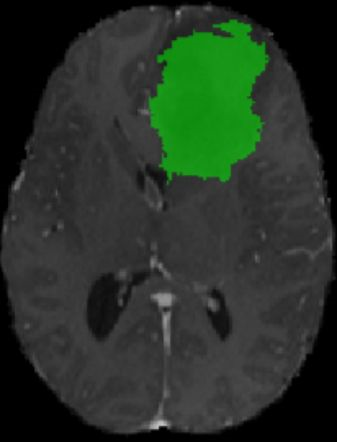}
     \includegraphics[height=\hhsize,width=\wsize\linewidth]{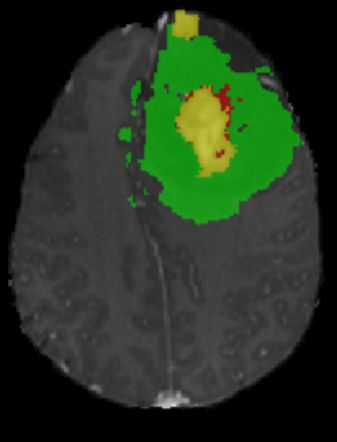}
 \end{minipage}
 \centering
   \caption{Result visualization from the BraTS training set (2020). Three axial slices of an unique sample in T1ce, corresponding channels output of the 3D encoder, corresponding channels input of the 3D decoder and corresponding labels. With Necrotic and Non-Enhancing Tumor core in red, Gadolinium-enhancing tumor in blue and peritumoral edema in green}
   \label{fig:result4}
\end{figure}

\section{Conclusion}

In this paper, we introduced a generic 3D U-Net architecture that
allows performance transfer by re-using and embedding any 2D
classifier network. The encoder as well as the decoder are composed
of two stages. The 3D input data goes through a process of depth
shrinking in order to transform the 3D data into 2D data. This
process is a succession of blocks of 3D convolutions and maxpooling
reducing the third dimension only. The transformed output data can
be then encoded by any 2D classification network. Moreover, decoding
also goes through a 2D decoding phase followed by a 3D decoding
procedure. Because of the limited computational resources, we
resized the images and trained separately the fourth modalities using four modality-specific networks. Nevertheless, the preliminary results seem to be promising.

Our goal was not to surpass all the BraTS sophisticated
segmentation techniques, but to provide a functional way to
re-use 2D classification architectures for 3D medical image
segmentation purposes. As can be seen, the learning transfer from weights trained on 2D natural images can be exploited for processing
3D medical images. We are convinced that we can significantly
improve the results by robustifying the learning technique, keeping
the original size of the data and stacking the 4 modalities
together.

\subsubsection*{Acknowledgements}
This work has been sponsored by the General Directorate for
Scientific Research and Technological Development, Ministry of
Higher Education and Scientific Research (DGRSDT), Algeria.

%
%
%
 \bibliographystyle{plain}
 \bibliography{BraTS_30}
%
%
\end{document}